\documentclass[conference]{IEEEtran}
\IEEEoverridecommandlockouts
\usepackage{cite}
\usepackage{amsmath,amssymb,amsfonts}
\usepackage{algorithmic}
\usepackage{graphicx}
\usepackage{textcomp}
\usepackage[table]{ xcolor }
\usepackage{times}
\usepackage{epsfig}
\usepackage{multirow}
\usepackage{amsmath}
\usepackage{amssymb}
\usepackage[table]{xcolor}
\usepackage{cite}

\def\BibTeX{{\rm B\kern-.05em{\sc i\kern-.025em b}\kern-.08em
    T\kern-.1667em\lower.7ex\hbox{E}\kern-.125emX}}
\begin{document}

\title{Smart IoT Cameras for Crowd Analysis based on augmentation for automatic pedestrian detection, simulation and annotation}

\author{\IEEEauthorblockN{1\textsuperscript{st} Antoine Rimboux}
\IEEEauthorblockA{\textit{Science, Engineering and Computing} \\
\textit{Kingston University}\\
London, UK \\
A.Rimboux@kingston.ac.uk}
\and
\IEEEauthorblockN{2\textsuperscript{nd} Rob Dupre}
\IEEEauthorblockA{\textit{Science, Engineering and Computing} \\
\textit{Kingston University}\\
London, UK  \\
R.Dupre@kingston.ac.uk}
\and
%\IEEEauthorblockN{3\textsuperscript{rd} Eldriona Daci}
%\IEEEauthorblockA{\textit{Department of Computing} \\
%\textit{Link Campus University}\\
%Rome, Italy \\
%e.daci@unilink.it}
%\and
\IEEEauthorblockN{3\textsuperscript{rd} Thomas Lagkas}
\IEEEauthorblockA{\textit{University of Sheffield International Faculty} \\
\textit{CITY College}\\
Thessaloniki, Greece \\
tlagkas@citycollege.sheffield.eu }
\and
\IEEEauthorblockN{4\textsuperscript{th} Panagiotis Sarigiannidis}
\IEEEauthorblockA{\textit{University of Western Macedonia} \\
\textit{Informatics and Telecommunications Engineering}\\
Kozani, Greece \\
psarigiannidis@uowm.gr}
\and
\IEEEauthorblockN{5\textsuperscript{th} Paolo Remagnino}
\IEEEauthorblockA{\textit{Science, Engineering and Computing} \\
\textit{Kingston University}\\
London, UK  \\
P.Remagnino@kingston.ac.uk}
\and
\IEEEauthorblockN{6\textsuperscript{th} Vasileios Argyriou}
\IEEEauthorblockA{\textit{Science, Engineering and Computing} \\
\textit{Kingston University}\\
London, UK  \\
Vasileios.Argyriou@kingston.ac.uk}
}

\maketitle

\begin{abstract}
Smart video sensors for applications related to surveillance and security are IOT-based as they use Internet for various purposes. Such applications include crowd behaviour monitoring and advanced decision support systems operating and transmitting information over internet. The analysis of crowd and pedestrian behaviour is an important task for smart IoT cameras and in particular video processing. In order to provide related behavioural models, simulation and tracking approaches have been considered in the literature. In both cases ground truth is essential to train deep models and provide a meaningful quantitative evaluation. We propose a framework for crowd simulation and automatic data generation and annotation that supports multiple cameras and multiple targets. The proposed approach is based on synthetically generated human agents, augmented frames and compositing techniques combined with path finding and planning methods. A number of popular crowd and pedestrian data sets were used to validate the model, and scenarios related to annotation and simulation were considered.
\end{abstract}


\section{Introduction}

IoT technologies incorporate a vast amount of specialized protocols and schemes \cite{triantafyllounetwork2018}, which have allowed them to enter a variety of domains. On this basis, they are eventually expected to dominate in a plethora of use cases, including very promising ones such as aerial communications for UAV-based monitoring applications \cite{lagkasuav2018}, where pervasive collection, annotation and processing of data is required, realizing the vision of ubiquitous computing. In the same time, multimedia traffic exhibits a continuous growth which is attributed to the evolution of computing devices (especially mobile devices), the increasing demands from the multimedia users' side, the improvements of video quality and content variety, as well as the significant developments in the telecommunications and networking infrastructure. Contemporary network multimedia services are now enabled through modern cyber-physical systems and can be provided through IoT autonomous distributed architectures utilizing agent-based middleware solutions \cite{eleftherakisarchitecting2015}. Furthermore, video streams (e.g. for surveillance) generated by IoT devices can now be routed over ad hoc wireless links and/or the fronthaul domain of the telecommunication system utilizing the flexibility provided by Software Defined Networking (SDN) and advanced techniques for optimized federated management over heterogeneous networks \cite{Bellavista2018,8332084}.

In the present video surveillance and situation security assume a certain level of understanding regarding the environment and the human or crowd behaviour \cite{Bloom01,Bloom02,Bloom03}. IoT applications such as surveillance, security, and market analysis require accurate crowd and pedestrian detection and behaviour recognition. Video processing and computer vision approaches based on deep {\em convolutional neural network} (CNN) architectures provide accurate estimates \cite{ZhangLLH16,2018arXiv180402047O}, but their accuracy depends on the quality and the amount of annotations in the training data sets. In order to improve the overall detection and recognition accuracy, a significant amount of labelled data is required, while the annotation process is time consuming, tedious, cost inefficient, prone to error, and often leads to performance degradation. Furthermore, the problem of how to perform comparative studies to simulation algorithms is considered in this work. The lack of a single and unified form of comparison between different simulation, detection and modelling approaches is still an issue for the crowd and pedestrian simulation and analysis methods. This often means that a given methodology is developed and evaluated for a specific purpose, with its wider abilities and properties left unconfirmed. Generally, the employed evaluation techniques and related measures can be broadly split into qualitative \cite{Portz2010} and quantitative  \cite{Kim2012, Asano2009}. The former including assessment made by experts in the field or context of the intended application \cite{Klugl2009}, as well as category based rating systems \cite{Duives2013} designed to define the capabilities of an algorithm.

%Pedestrian and crowd simulation has applications in a wide range of industries including pedestrian facility suitability and capacity \cite{Asano2009}, computer graphics and gaming \cite{Kim2012}, the social sciences \cite{Klugl2009} and engineering \cite{Xi2011}. This broad range of uses has led to extensive research into how crowds and pedestrians move around and interact with their environment.
%
A number of quantitative measures have been suggested to provide a numeric measure of accuracy for detection and simulation, to mention a few: bounding boxes, tracks, speed, pedestrian density, number of steps taken to destination and duration. These evaluation techniques tend to be data driven, and as such require ground truth data for testing purposes. The concept of an evaluation framework has been suggested before \cite{Charalambous2014, Wolinski2014, Guy2012, Kapadia2011, Rodreiguez2011, Musse2012}; with most deducing various metrics based on a simulation in an effort to rate simulation algorithms or tune parameters. Many of these evaluation frameworks are affected by problems tightly related with the data collection and annotation process, such as cost, time, privacy and suitability and availability of large outdoor environments.

With the advent of so much research in the area of pedestrian detection and crowd analysis, the ability to generate realistic data of different modalities, including ground truth has been researched over the last few years. It is worth mentioning the work of \cite{7299006, 7780721, Handa2016, Duives2013, Papadimitriou2009, singh2009, Zhan2008,Kim2012,Portz2010,Pettre2009} on data generation and simulation algorithms for crowd and pedestrian analysis.

In \cite{Charalambous2014}  the creation of a tool is proposed that characterises and generates outlying behaviour in simulated videos. In \cite{Guy2012} an entropy score is computed to generate simulated data closer to real world data.  The {\em stochastic variational dual hierarchical Dirichlet process} (SV-DHDP) model is introduced in  \cite{Wang2016}, where groups of similar trajectories (trending paths) and subpixel motion flows \cite{Argyriou03, Argyriou04, Argyriou05} can be combined to generate an overall path pattern for an environment offering higher realism. In \cite{Lerner2009, Lerner2010}  the concept of {\em look and feel} of a crowd is proposed by comparing an agent's actions at a given moment in time with a database of observed actions, providing more realistic simulated videos. In \cite{Musse2012} the issue of tracking generalised paths in crowds is tackled using four dimensional histograms to describe and generate movement within a crowd. Additionally the work in \cite{Rodreiguez2011, Banerjee2011,Kapadia2011,Zanlungo2014} propose interesting approaches and metrics to generate realistic crowd behaviours.

The use of synthetic data generation to train and evaluate machine learning models based on deep architectures was introduced recently. Methods in the literature use the process to generate labelled data sets for different applications, including pedestrian and crowd analysis. These methods aim to generate pedestrians and groups of people at different locations in a given scene, supporting a variety of appearances. However, current methods are either restricted to a single camera in terms of their usage or support only single frames without motion to be considered. Additionally, all of them are based on 3D virtual environments with quality not comparable with real video sequences affecting the obtained models and under-performing in real scenes. In this paper the proposed framework supports multiple cameras and moving agents, generating videos instead of single images based on compositing, a technique used to generate realistic videos by superimposing virtual object in real scenes. As such, the following novel framework is suggested which reduces the complexity of crowd and pedestrian realistic simulation, allows the automatic generation of data sets and annotation, supporting different data modalities, aiming to improve the detection accuracy of existing deep architecture focusing on pedestrian detection, pose estimation and tracking.
\section{Simulation Evaluation using Augmentation} \label{sec:5_Sim_Evaluation}

The proposed modular {\em crowd composition framework} (CCF) provides a method of pedestrian and crowd data augmentation which considers realistic simulated human walking behaviours, multi-view and multi-modal data. Data generation can be implemented on a frame by frame basis or for a sequence as a whole, providing flexibility on how simulated data and  ground truth are extracted. Additionally, the proposed methodology requires no track or path information, allowing the user to control parameters related to the number of pedestrians, their behaviour and the modality of the data.

\begin{figure}
    \centering
    \includegraphics[width=0.95\columnwidth]{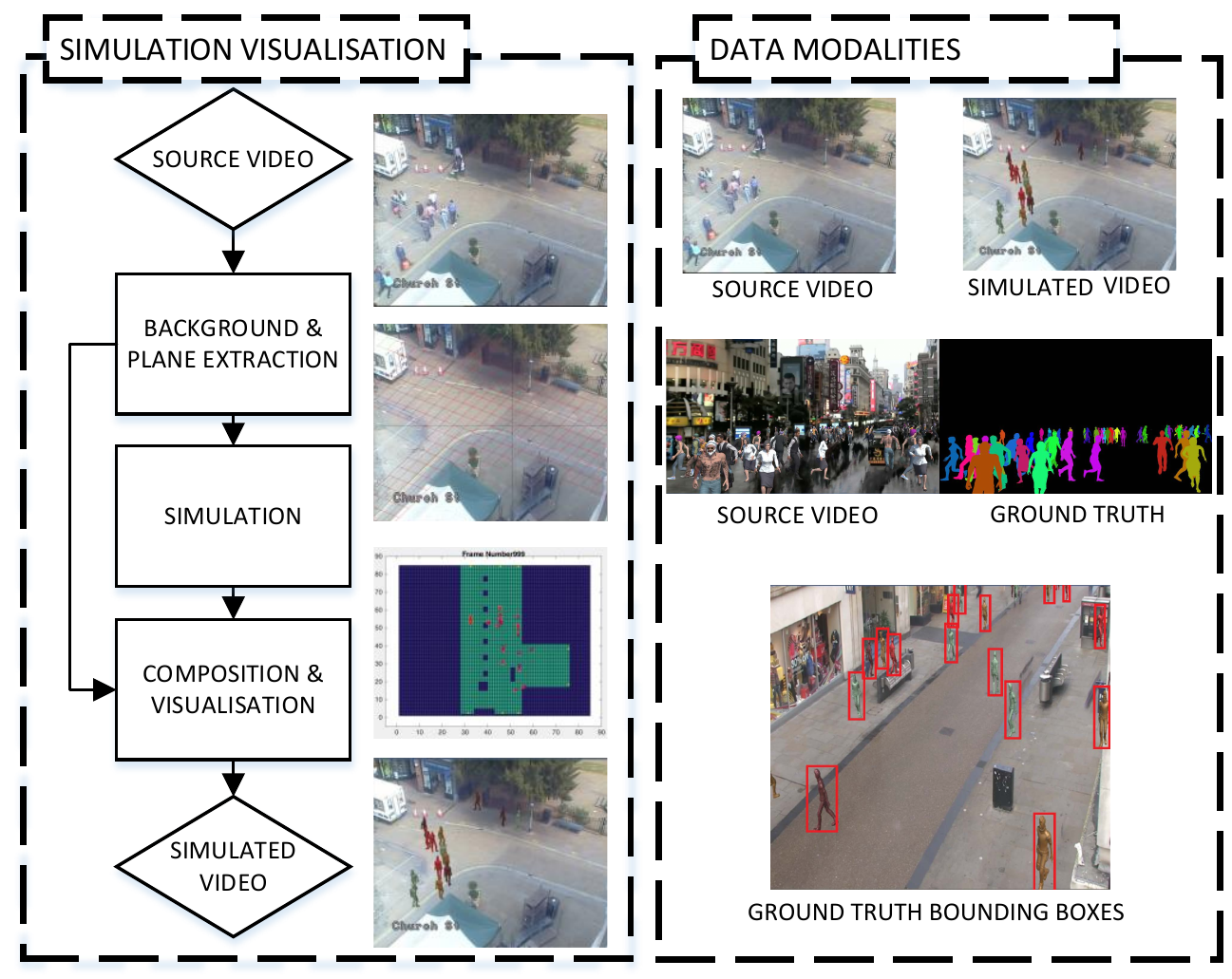}
    \caption{Overview of the CCF framework.}
    \label{fig:5_Overview}
\end{figure}

The proposed framework takes as input source video footage and generates an augmented output video using composition techniques. The process utilises background subtraction techniques as well as methods to extract 3D data from 2D images. This allows the construction of a 3D space in which virtual agents can navigate around. Through the use of composition, a final visualisation combining this background and 3D space is generated to form the simulated video sequence in which artificial agents are superimposed onto the background of the source video data.
%
%\begin{figure*}
%    \centering
%    \includegraphics[width=15cm]{Modular_Framework}
%    \caption{Outline of the modular nature of the Crowd Simulation Evaluation through Composition (CSEC) framework.}
%    \label{fig:5_Modular_Framework}
%\end{figure*}
%
Fundamentally the framework is made up of two components: simulation visualisation and annotated video data generation. The modular nature of the framework supports inputs of any simulation algorithm or video analysis techniques, depending on application. Furthermore, it retains the ability to produce realistic synthetic annotated crowd data. Figure \ref{fig:5_Overview} provides a more specific overview of the CCF framework as it is utilised in this work.
%
%\begin{figure}
%    \centering
%    \includegraphics[width=14cm]{FIGURES/OVERVIEW}
%    \caption{Overview of the Crowd Simulation Evaluation through Composition (CSEC) framework.}
%\end{figure}
%
%\begin{figure}
%    \centering
%    \includegraphics[width=0.45\columnwidth]{KRAD2_SOURCE}
%    \includegraphics[width=0.45\columnwidth]{KRAD2_MEDIUM}
%    \caption{Frames of source CCTV footage and generated video using the composition techniques.}
%    \label{fig:5_Initial_Comparison}
%\end{figure}

As the proposed framework uses videos to generate synthetic data, firstly a simulated video must be constructed. Initially, using the source video sequence, the background is obtained. Next, a two dimensional plane is extracted representing a top down view of the given scene. Simulations are then run to produce paths for the virtual agents to follow based on the extracted plane. The visualisation component is then used to create a composite of the extracted 2D background image and 3D rendered agents as they follow the simulated paths. Frames are output from the visualisation into a final simulated video sequence. Once both a simulated and source video are available, the ground truth and other data modalities such as depth can be exported. Finally, {\em tracklets}, pose, skeleton, flow, and density measures can be evaluated and used for training and evaluation in deep architectures or other learning techniques.

\begin{figure}[!ht]
    \centering
    \includegraphics[width=0.6\columnwidth]{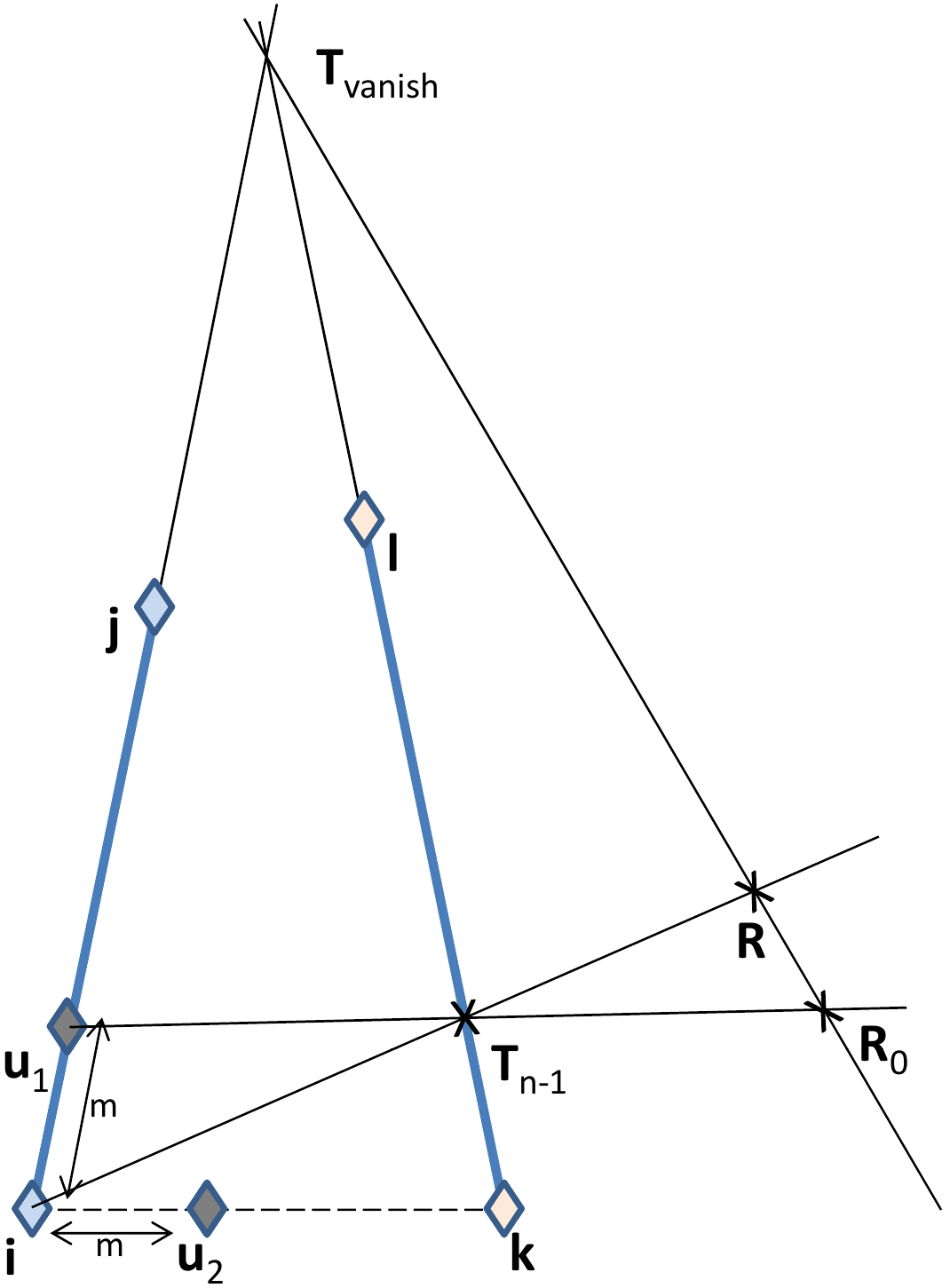}\\ \vspace{3mm}
    \includegraphics[width=0.6\columnwidth]{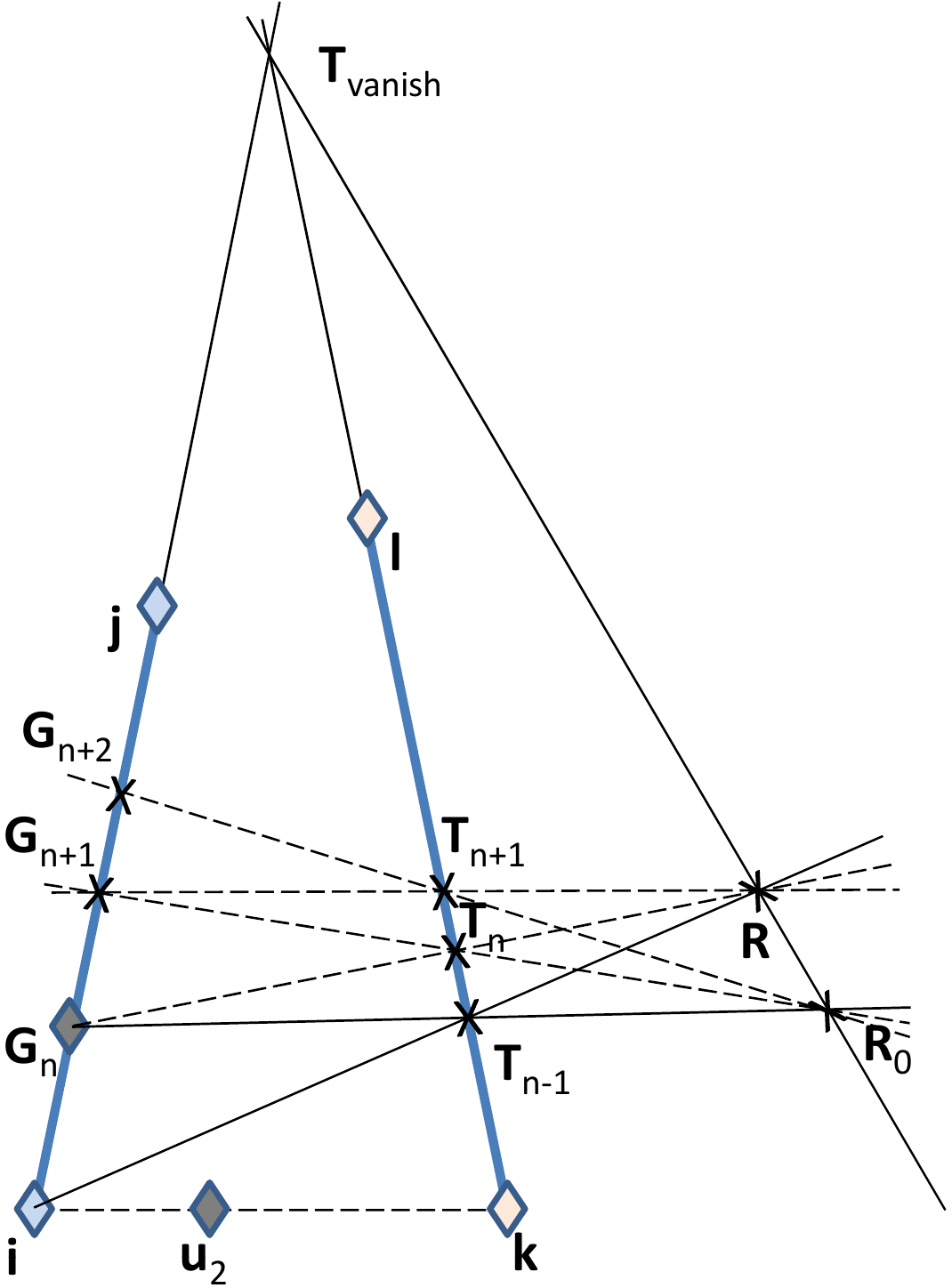}
    \caption{(a) User defined points and initialisation. (b) The first two iterations of the recursive algorithm. }
    \label{fig:PerspectiveGrid}
\end{figure}

To allow the composition of the simulated video to be created, the background of the source video sequence is required \cite{Zivkovic2004}. Once the background image has been subtracted the process of defining the perspective grid is applied. The perspective grid allows scale mapping of an environment from the viewpoint of the source video camera pose. The resultant grid represents a top down environment map of the viewable area and is used during agent simulation. Using the concept of perspective scale along a line we can, through the definition of two parallel lines that run to the vanishing point of an image, estimate distance in arbitrary units of measure within this perspective space (Figure \ref{fig:PerspectiveGrid}b). This unit can be based upon an object in the scene with known dimensions or using pedestrians \cite{Chan2008}.

Initially the user defines the points $\mathbf{i}$ and $\mathbf{j}$, in the 2D image space, forming a line along an edge that leads to the vanishing point of the image. A second line is defined by the points $\mathbf{k}$ and $\mathbf{l}$, such that it runs {\em parallel}, relative to the vanishing point in the 3D space of the captured image, to the line defined by points $\mathbf{i}$ and $\mathbf{j}$ (Figure \ref{fig:PerspectiveGrid}a). At a location along the line $\mathbf{i} \mathbf{j}$ the user defines another point $\mathbf{u}_1$, such that the line $\mathbf{i} \mathbf{u}_1$ represents the unit of distance $m$ from which all further perspective points are defined. An additional point $\mathbf{u}_2$ is defined on top of the line $\mathbf{i} \mathbf{k}$ which represents the same relative distance in 3D space as $m$. For the next step of the proposed algorithm the reference points $\mathbf{T}_{vanish}$, $\mathbf{R}$, $\mathbf{R}_0$ and $\mathbf{T}_{n-1}$ are initialised automatically (Figure \ref{fig:PerspectiveGrid}a).
In more detail, the vanishing point $\mathbf{T}_{vanish}$ is defined as the point at which the lines $\mathbf{i} \mathbf{j}$ and $\mathbf{k} \mathbf{l}$ intersect, this may well be at a position outside the image plane. As such $\mathbf{T}_{vanish}$ is defined as
\begin{align}
    \mathbf{T}_{vanish} = f(\mathbf{i}, \mathbf{j},  \mathbf{k}, \mathbf{l})
\end{align}
An arbitrary point $\mathbf{R}$ is selected at a random location outside the triangle $\mathbf{i} \mathbf{T}_{vanish} \mathbf{k}$. The point $\mathbf{T}_{n-1}$ is defined as the point of intersection of the lines $\mathbf{i} \mathbf{R}$ and $\mathbf{k} \mathbf{T}_{vanish}$
\begin{align}
    \mathbf{T}_{n-1} = f(\mathbf{i}, \mathbf{R},  \mathbf{k}, \mathbf{T}_{vanish})
\end{align}
Finally the point $\mathbf{R}_0$ is defined.
\begin{align}
    \mathbf{R}_0 = f(\mathbf{u}_1, \mathbf{T}_{n-1}, \mathbf{R}, \mathbf{T}_{vanish})
\end{align}

%\begin{figure}[!ht]
%    \centering
%    \begin{subfigure}{.48\textwidth}
%        \centering
%        \includegraphics[width=7.4cm]{FIGURES/Perspective1}
%        \caption{}
%    \end{subfigure}
%    \begin{subfigure}{.48\textwidth}
%        \centering
%        \includegraphics[width=7.4cm]{FIGURES/Perspective2}
%        \caption{}
%    \end{subfigure}
%    \caption{(a) User defined points and initialisation. (b) The first two iterations of the recursive algorithm. }
%\end{figure}

With these points initialised, a recursive algorithm is applied to calculate equidistant points along the line $\mathbf{i} \mathbf{T}_{vanish}$ in 3D space. As the user has already defined the first of these points $\mathbf{u}_1$, for the purposes of the recursive step, these will be relabeled as $\mathbf{G}_n$.
This is a two-step iterative process, with the point $\mathbf{T}_n$ being defined as the intersection of the lines $\mathbf{G}_n \mathbf{R}$ and $\mathbf{k} \mathbf{T}_{vanish}$.
\begin{align}
    \mathbf{T}_n = f(\mathbf{G}_n, \mathbf{R}, \mathbf{k},\mathbf{T}_{vanish})
\end{align}
and during the second step the next equidistant point $\mathbf{G}_{n+1}$  on the line $\mathbf{i} \mathbf{T}_{vanish}$ is expressed as a function of
\begin{align}
    \mathbf{G}_{n+1} = f(\mathbf{R}_0, \mathbf{T}_n,  \mathbf{i}, \mathbf{T}_{vanish})
\end{align}
This process is repeated until $\mathbf{G}_{n+1}$ is no longer within the borders of the original background image. The grid is initially defined using all the equidistant points on the line $\mathbf{i} \mathbf{k}$, using the distance $\mathbf{i} \mathbf{u}_2$ as a unit. Lines are defined between each of these points and the vanishing point $\mathbf{T}_{vanish}$ of the image. The scale points $\mathbf{G}$ are plotted along each of these newly defined lines forming the grid. Additionally, if required, the recursive process can be inverted to create points moving away from the vanishing point. This ensures that the entire image plane is encapsulated by the defined grid, regardless of where the user has defined their points. The resultant grid represents the perspective plane of the source image. On that grid the areas (cells) with obstacles (i.e. cells where pedestrians cannot walk) are annotated as is information about entrance/exit locations.
\begin{figure}
    \centering
    \includegraphics[width=0.85\columnwidth]{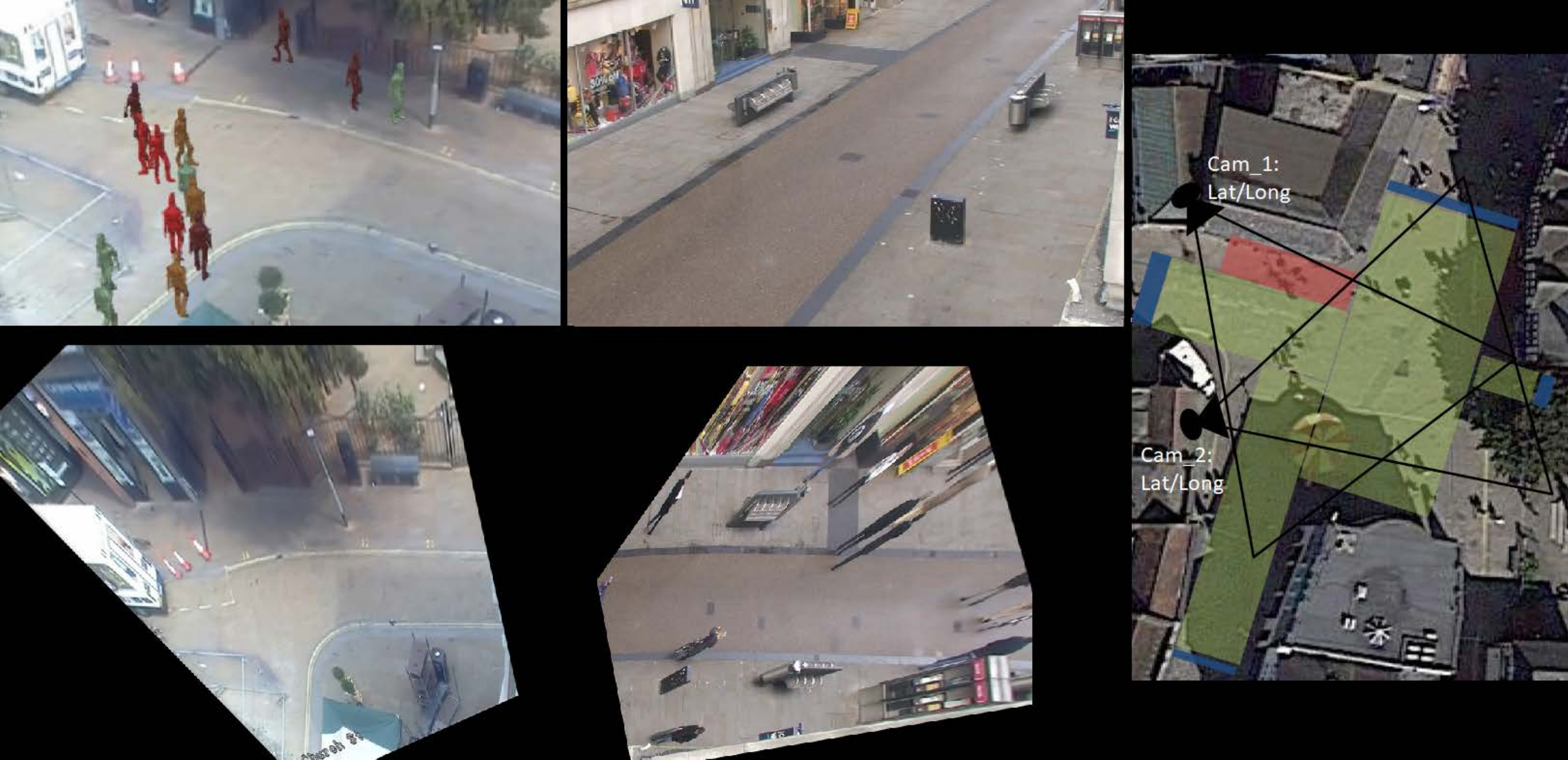}
    \caption{An example of multiple camera simulation and the corresponding projections.}
    \label{fig:MultiCam}
\end{figure}
By adding a few additional calibration variables, during the plane extraction phase, the composition and visualisation process can be extended to multi camera setup steps, both in overlapping and non-overlapping scenes (Figure\ref{fig:MultiCam}). In these cases, rather than using a single virtual camera within a composition scene, additional cameras are added to represent the other camera views. During the calibration of these cameras, the same plane extraction technique is used, however additional relational measurements (latitude/longitude and orientation or distance and bearing from the source camera) are required to allow the positioning of these additional cameras within the 3D environment. The same process of background image alignment is also used on the additional virtual cameras to allow the generation of composite videos from these new views. The principle advantage here is that now a single crowd simulation can be viewed from several different camera angles whilst still having access to the ground truth data and crowd statistics.

\section{Evaluation of the Crowd Composition Framework}

\begin{figure}
    \centering
    \includegraphics[width=0.85\columnwidth]{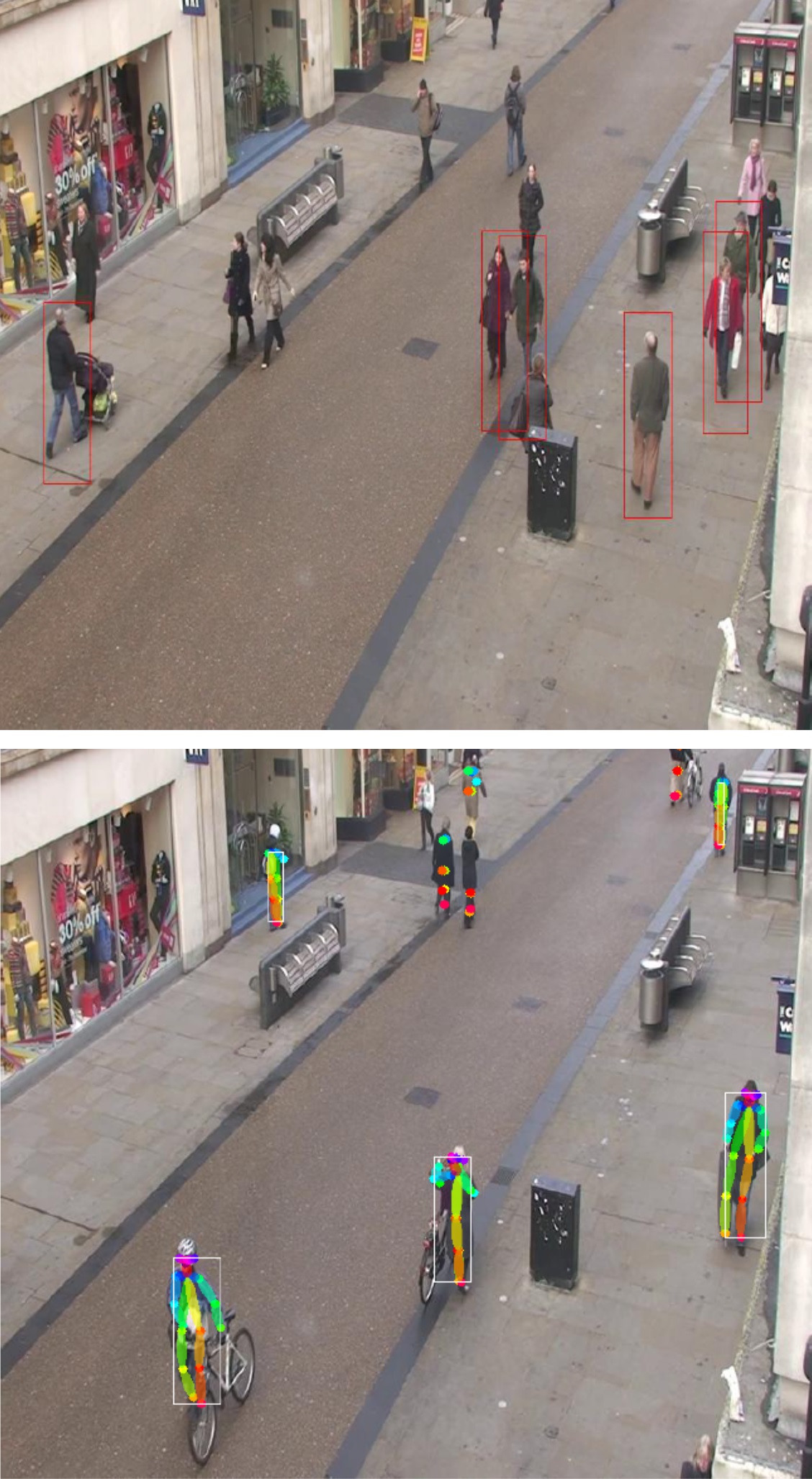}
    \caption{Left: Bounding boxes estimated by the RPN network. Right: Skeleton and the corresponding bounding boxes detected by the VGG-19 network.}
    \label{fig:Stud01}
\end{figure}

This part of the paper is focused on the evaluation of the proposed CCF framework using a set of different deep neural networks trained for applications related to pedestrian detection. The first model used was \cite{8099979} and is a {\em region proposal network} (RPN), used for the detection of pedestrians in abnormal situations. This model takes as input images of size $960\times720$ and, returns bounding boxes for all pedestrians present in the images (see figure \ref{fig:Stud01} left). The second model proposed in \cite{cao2017realtime} is based on the ResNet-101 deep network and aims to estimate the pose of different persons on a image. This network was trained using the COCO data set \cite{502} where annotations are not made of boxes but keypoints which are, for pedestrians, their joints (such as knee, ankle and neck) as we can see in figure \ref{fig:Stud01} right.

\begin{figure}
    \centering
    \includegraphics[width=0.85\columnwidth]{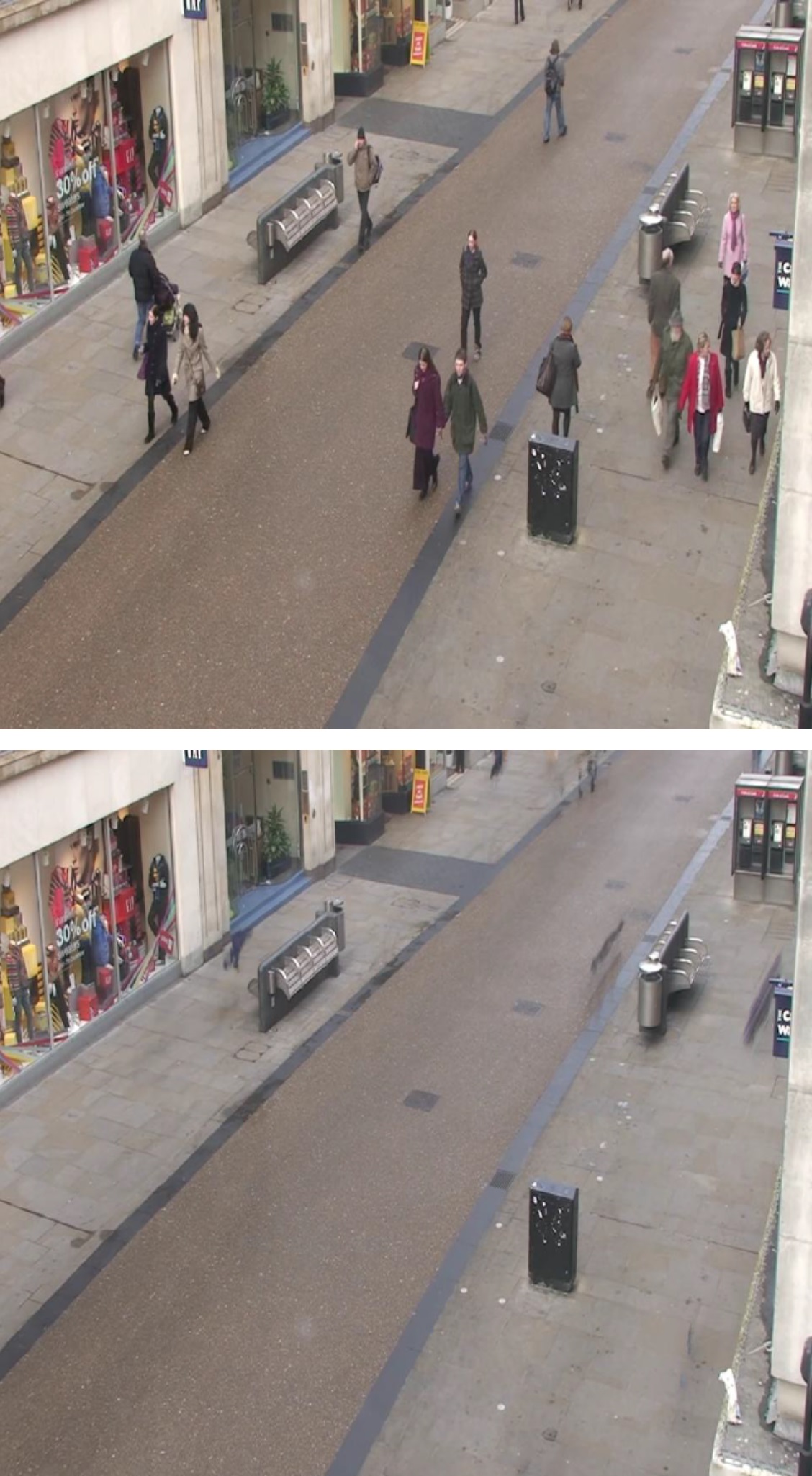}
    \caption{Left: A frame from the `Town Centre' video dataset. Right: Extracted Background.}
    \label{fig:Stud02}
\end{figure}

In order to evaluate the contribution of the proposed CCF framework these networks are retrained using additional simulated data. Therefore, the data set that was used for the training and the comparison with the synthetic data is the {\em Town Centre} data set \cite{benfold2011stable} (see figure \ref{fig:Stud02}) which represents the footage of a crowded street captured by a CCTV camera. This video is supplied with an annotation file which contains, the coordinates of the boxes bounding for the pedestrians present for each frame. Note that this data set will also be used as a reference to build the synthetic data. As such the background is extracted based on the proposed framework and is used int he composition of the synthetic data (see figure \ref{fig:Stud02} right). Also, the synthetic data are designed to be similar to real life data so there will be a need to analyze the path of the pedestrians in the video to make the simulated pedestrians follow similar paths (e.g. select same entrance and exit points in the scene).

\subsection{Evaluation Metrics}

In order to evaluate the accuracy of the models, metrics must be defined to determine success of failure in the detection of a pedestrian. Since the provided data set provides annotation files, the ground truth is provided as bounding boxes. Additionally, as it was previously stated, when an image is input, the model returns several boxes and the corresponding scores (which can be viewed as the probability for the region of the image located inside the box to actually represent a pedestrian). If the output is not formatted like so, for instance as the pose estimation model \cite{cao2017realtime}, estimation of the corresponding bounding boxes is performed. In this particular case, the top left corner of the box is defined by the lowest $x$ and $y$-coordinates of the human joints. The bottom right corner is selected based on the highest $x$ and $y$-coordinates.

Thus, a means to compute the accuracy of the model would be to compare, for a given image, the ground-truth boxes from the annotation file with the highest scored boxes that have been returned by the model for the same image. For this work {\em intersection over union} (IoU) is used to determine the accuracy of the selected deep networks trained with and without the simulated data. So for two boxes, the ground truth and the prediction, computing this overlap consists of dividing the area of intersection between the boxes by the area of union. With $A$ and $B$ the areas of the two boxes we have
\begin{align}
    IoU = \frac{A\cap B}{A\cup B} \cdot 100
\end{align}
To compute the global accuracy, each predicted box with a score higher than a given threshold (most likely to represent a pedestrian) is compared with their closest ground-truth box (using the {\em Manhattan} distance). The overlap percentage for this ground-truth box is then added to the annotation file. Therefore, this file is made of the ground-truth data and the overlap for each box, which represents the accuracy of the box predicted by the model. Moreover, if a ground-truth box does not have an added overlap, that means that no predicted boxes where close enough to have a non-null union. In that case, the model did not manage to predict the location of the pedestrian represented by this ground-truth box. Also, if several predicted boxes are the closest to the same ground-truth box, the highest overlap is kept.

Finally, to compute the accuracy of the model in an image, the last part is to sum the overlap for each ground-truth box and divide it by the number of pedestrians present in the image to obtain the average accuracy. If an overlap for a ground-truth box passes a given threshold (e.g. $80\%$) the pedestrian is considered as found.

\subsection{Crowd simulation process}
The next step of the evaluation process is to build a crowd simulation made of synthetic agents using the proposed CCF framework, capturing the video frames and automatically generated ground truth.

\begin{figure}
    \centering
    \includegraphics[width=0.75\columnwidth]{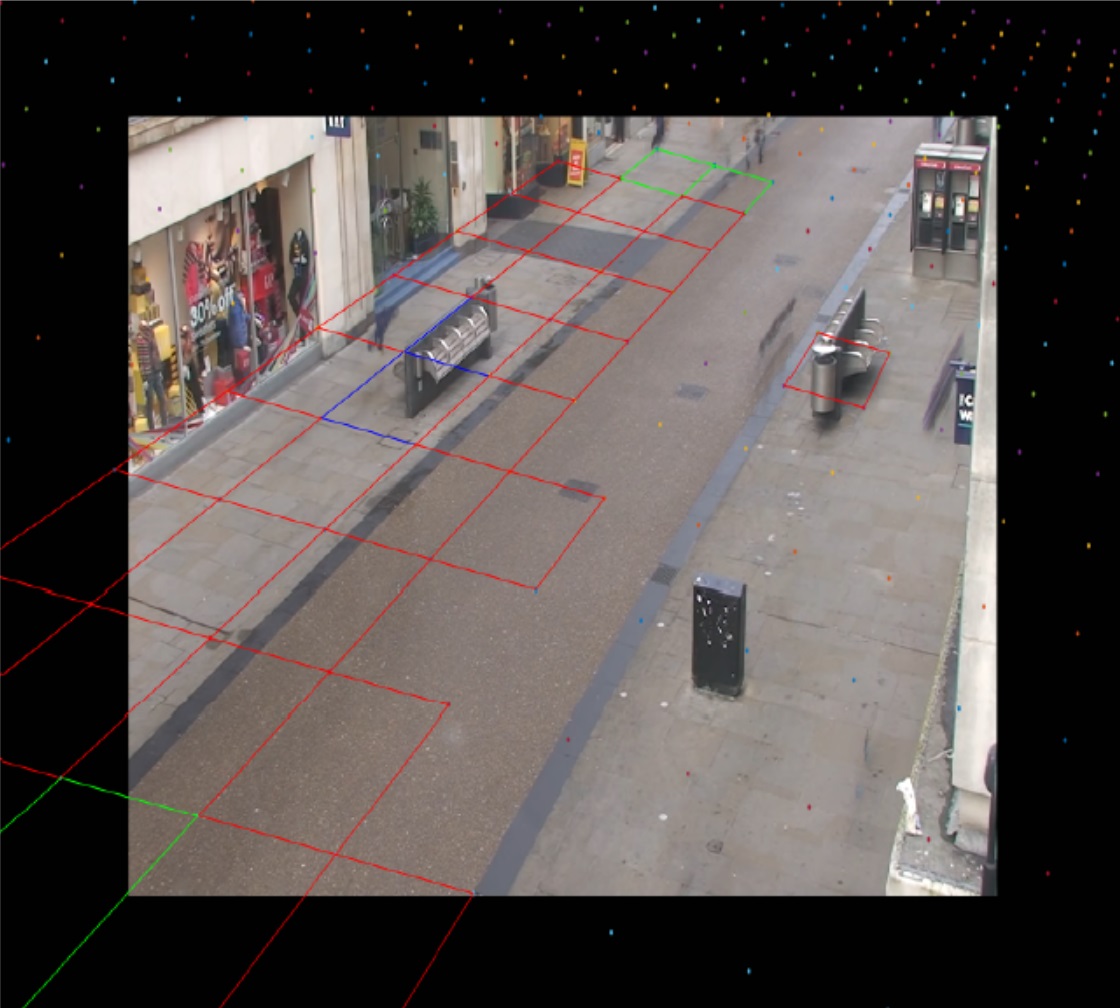}\\ \vspace{3mm}
    \includegraphics[width=0.75\columnwidth]{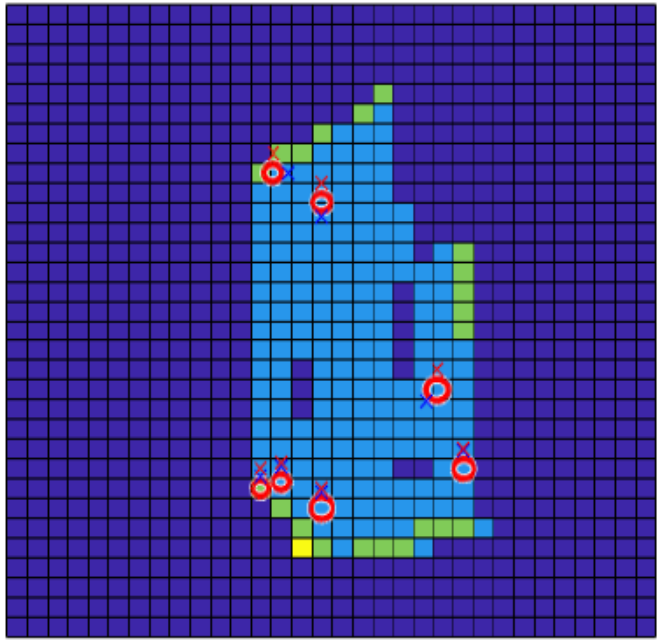}
    \caption{a) Computed grid, partially filled. b) Top down view of the computed grid and visualization of the agents’ paths.}
    \label{fig:Stud03}
\end{figure}

Firstly, using CCF we extract the background of a sequence of images or the whole video, compute a grid that represents the walkable area for the synthetic agents and, then provide the characteristics of the synthetic agents (e.g. height, speed, entrance and exit points). We estimate the path of each agent but also its dimensions and orientations according to the obtained perspective. In this particular case, using the {\em Town Centre} data set, the first step is to extract the background of the video as it was analysed above. Once the background is extracted, the next step is to build the grid (or map) that will represent the area on which the agents will walk, following the perspective of the scene. According to the CCF framework, we provide two parallel lines and a unit (i.e. a square of size $1\times 1$ meter) and the framework returns the map (see figure \ref{fig:Stud03}a).
All is left to do is to decide which part of the map is walkable, which parts are obstacles, and where the entrances and exits are situated. Then, with the details about the characteristic of the agents, the framework computes the path of each agent (see figure \ref{fig:Stud03}b) using a crowd simulation algorithm and returns their positions, orientation and scale information that is used for the scene simulation and the annotated data extraction.

\begin{figure}
    \centering
    \includegraphics[width=0.75\columnwidth]{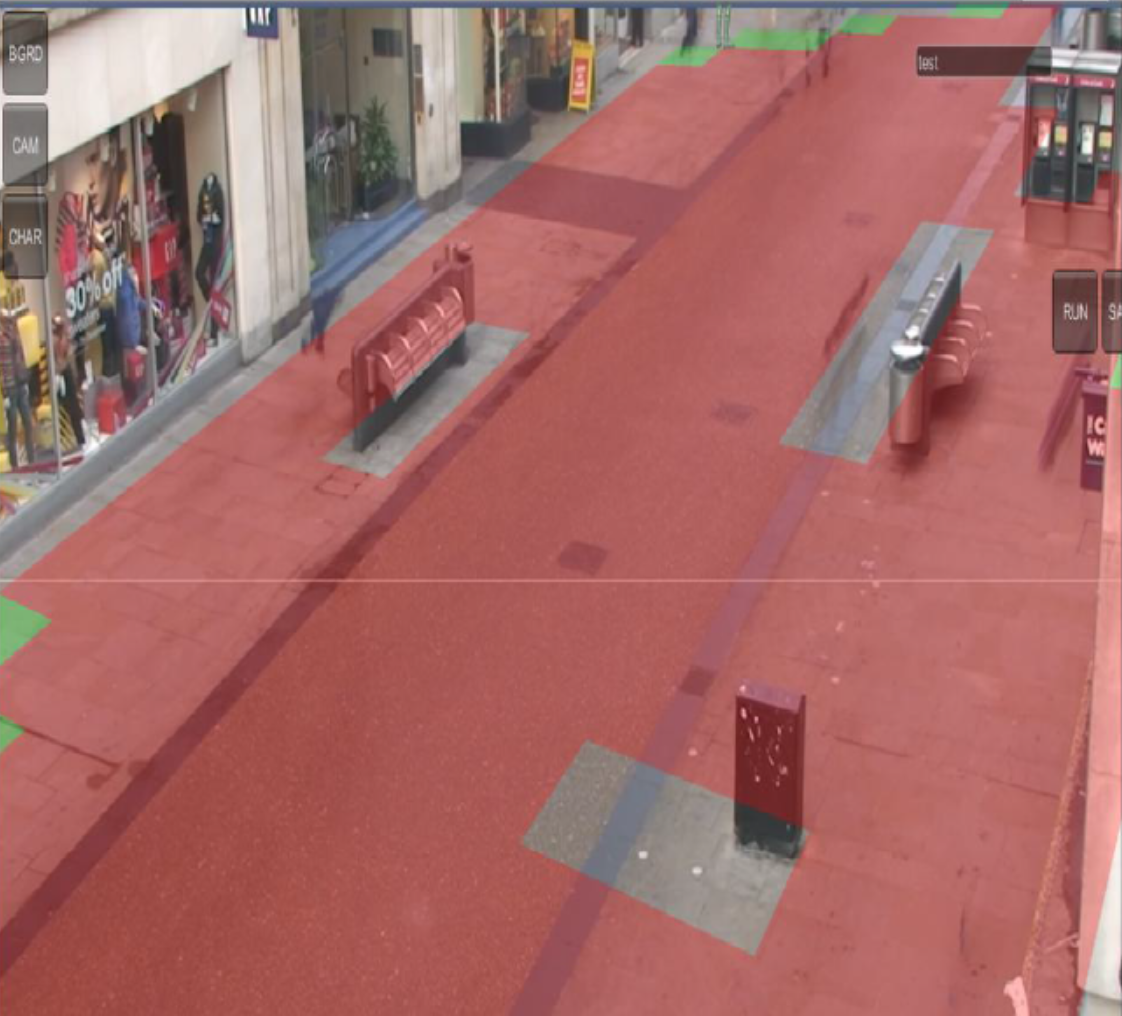}\\ \vspace{3mm}
    \includegraphics[width=0.75\columnwidth]{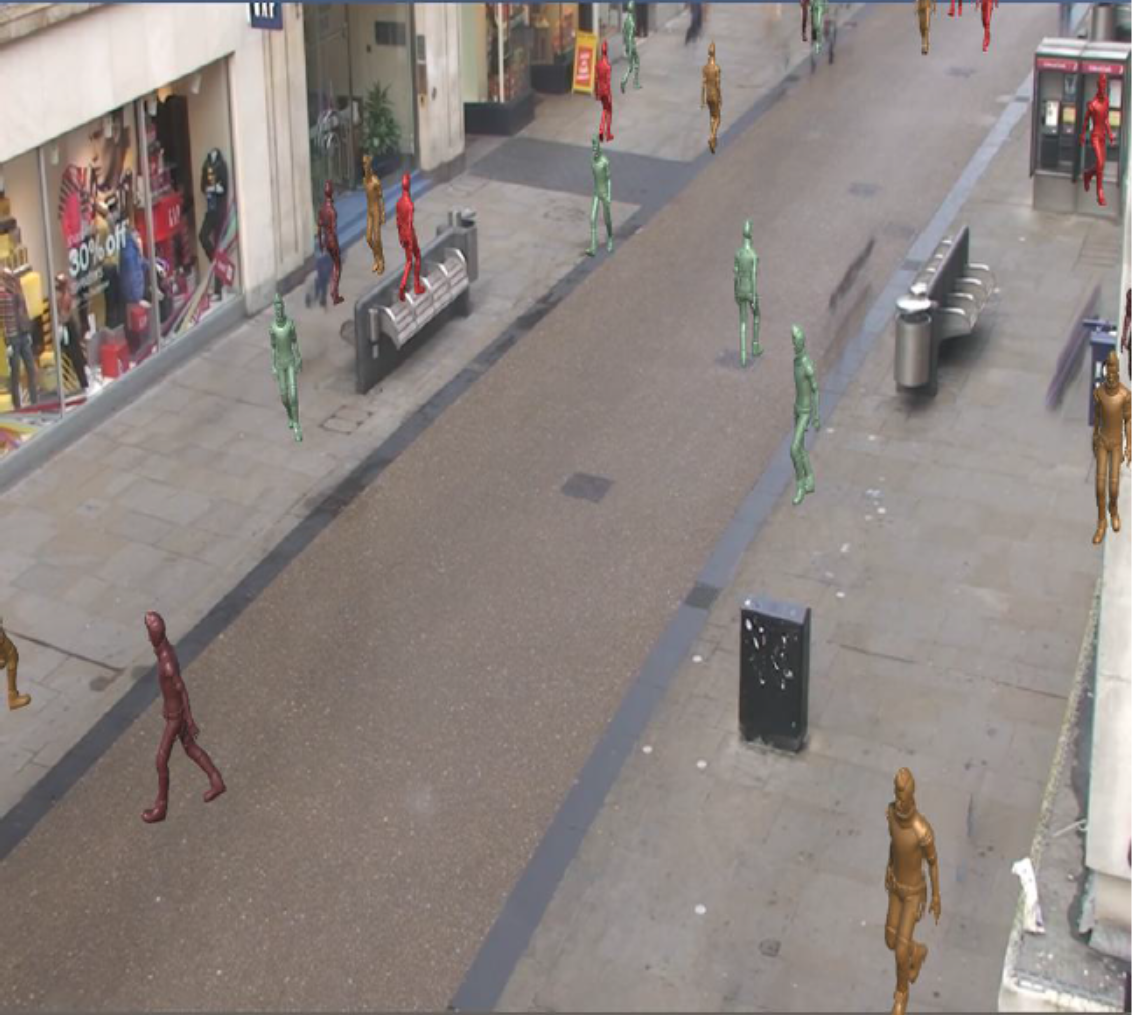}
    \caption{a) Superposed map on the extracted background. b) Crowd simulation composited frame.}
    \label{fig:Stud05}
\end{figure}

Note that, in order to provide an adequate amount of images for the training, approximately $100$ agents are selected. The entrance and exit points are assigned for each agent randomly but with weighted probabilities based on the observed behaviours in the real video. Examples of the superposed map on the extracted background showing the walkable areas and the obtained composited crowd simulation are shown in figure \ref{fig:Stud05}.
Next the simulation is run showing the agents walking on the map emulating a real crowd. Finally, for each rendered frame, the RGB image and the corresponding ground truth data are saved.
%
%\begin{figure}
%    \centering
%    \includegraphics[width=0.95\columnwidth]{Stud00}
%    \caption{Rendered frame with composition and the corresponding ground truth.}
%    \label{fig:Stud00}
%\end{figure}
%
 At each frame, the agents in the scene are rendered according to their different characteristics (coordinates, rotation) in the map coordinate space. So at each frame, the CCF framework provides the ground truth as an image and as a list of bounding boxes. The generated synthetic data set includes $2845$ images and it is used to retrain the available networks.

\subsection{Fine-tuning the selected networks}
The fine-tuning and retraining process for the selected RPN, and ResNet-101 networks, employed mainly to re-scale the input images and format the ground truth to the expected dimensions and order. For each network the original parameters were selected and the new simulated data added to obtain the new models.

\subsection{Results}
\begin{figure}
    \centering
    \includegraphics[width=0.75\columnwidth]{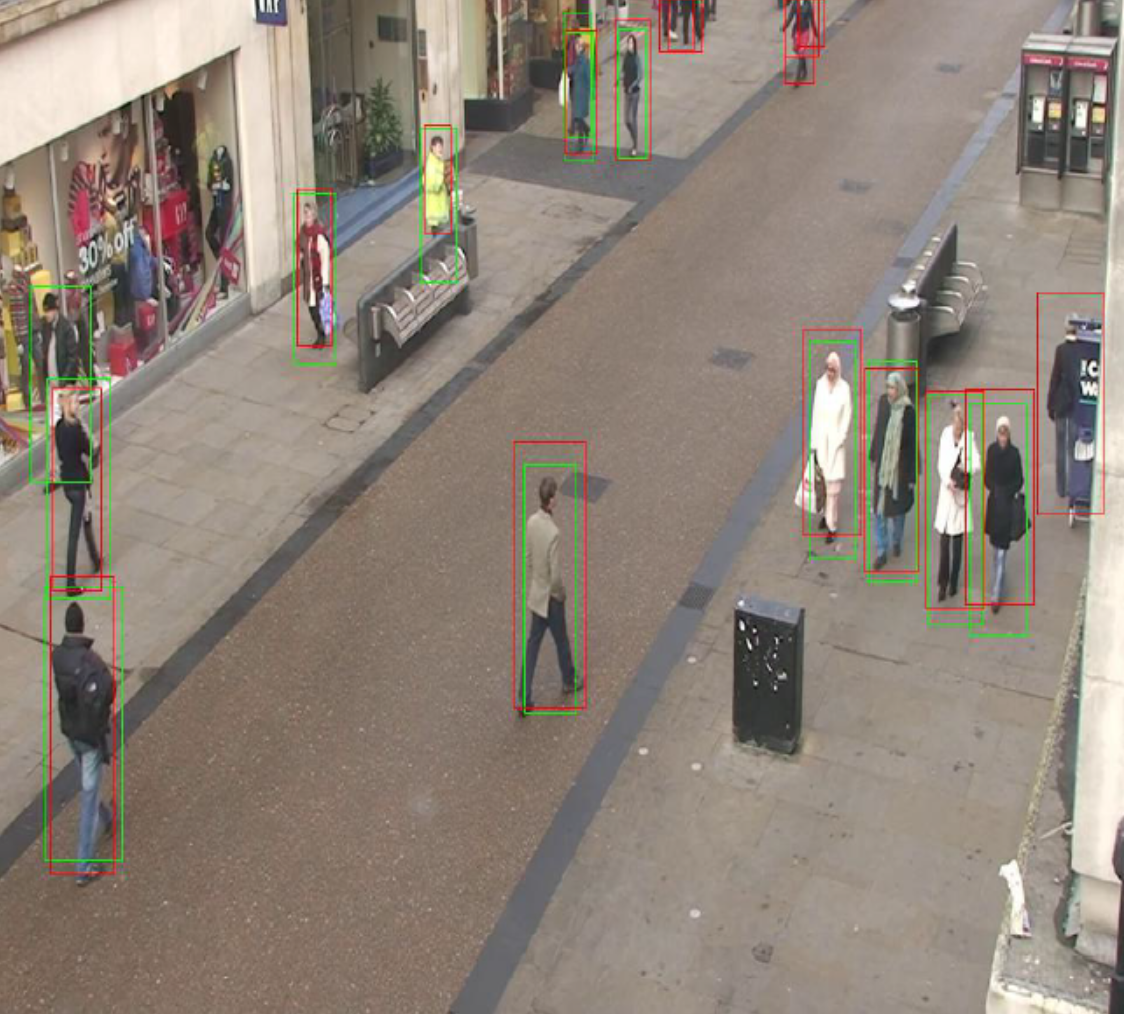}\\ \vspace{3mm}
    \includegraphics[width=0.75\columnwidth]{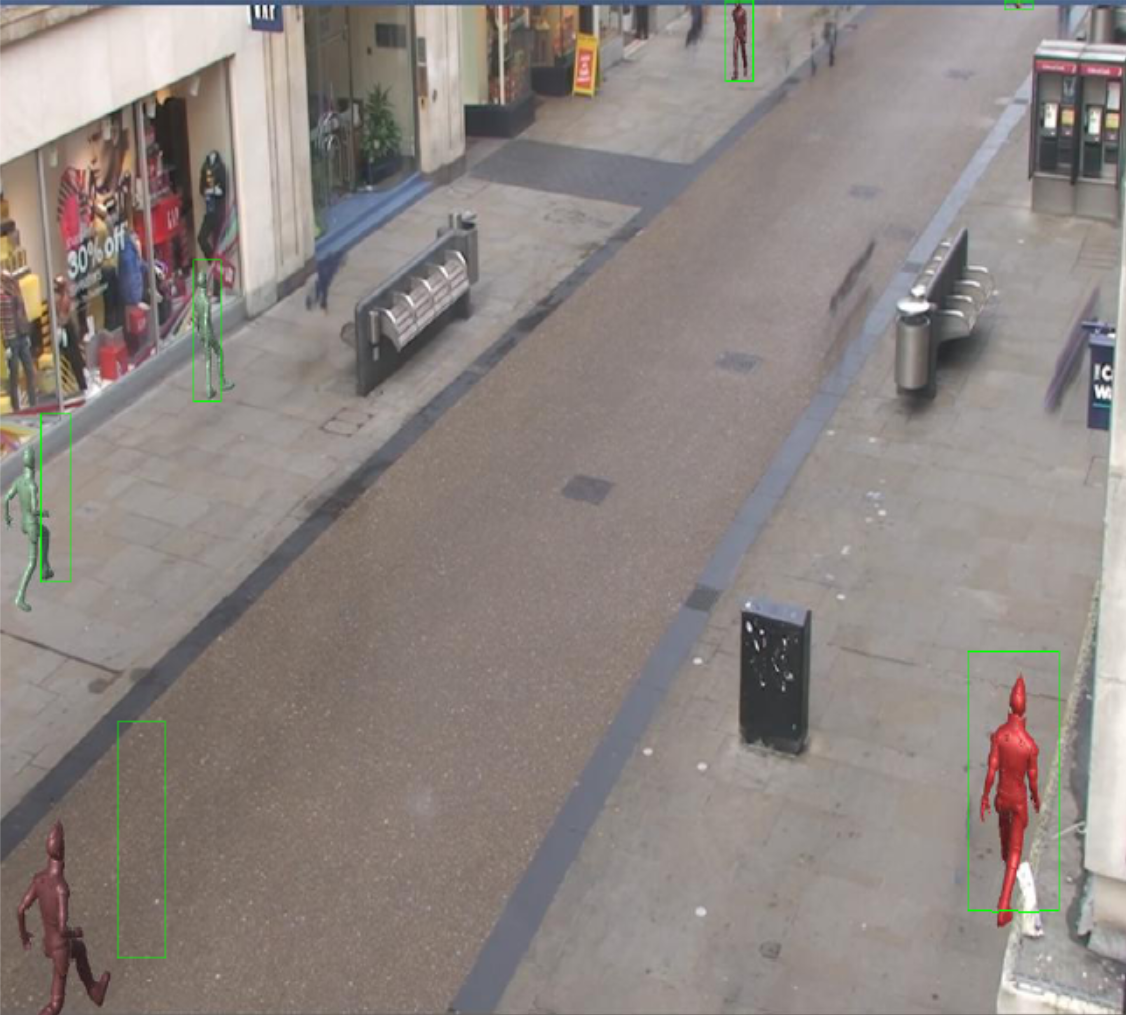}
    \caption{a) Results using the ResNet-101. b) Examples of False positives using the simulate data.}
    \label{fig:Stud07}
\end{figure}

Evaluation is carried out using the metrics previously presented. The table \ref{sample-table} shows the average accuracy. For the validation step, the read frames from the {\em Town Centre} data set were used. Results demonstrate that the synthetic data set generated using the proposed CCF framework can be utilised to train deep neural network and significantly improve their accuracy to detect pedestrians in real video sequences.
\begin{table}[t]
  \caption{Obtained accuracy for the selected deep networks with and without simulated data.}
  \begin{center}
    \begin{tabular}{c | c c}
      \hline
      \hline
      Model-Accuracy & Original & Proposed CCF    \\
      \hline
      ResNet-101 & 39.09\%   & \textbf{50.04}\% \\
      RPN+ & 20.77\%   & \textbf{21.32}\% \\
      \hline
      \hline
    \end{tabular}
    \label{sample-table}
  \end{center}
\end{table}
Qualitative results for both training and testing stages are shown in figure \ref{fig:Stud07}.

\section{Conclusions}
A novel {\em crowd composition framework} was presented which provides simulated annotated data using the composition process for pedestrian detection. Through the use of a modular system, any crowd or pedestrian simulation model or data annotation system supporting multiple cameras can be evaluated and compared by generating agent motion for use in the final visual simulation. Additionally, any video analysis feature can be utilised to evaluate similarity. Our experiments showed that the proposed framework improved the performance of deep networks in terms of pedestrian detection and crowd analysis.

\section*{Acknowledgment}

This work is co-funded by the NATO within the WITNESS project under grant agreement number G5437. The Titan X Pascal used for this research was donated by NVIDIA.

\bibliographystyle{IEEEtran}

\bibliography{THESIS}

\end{document}